\documentclass[11pt]{article}

\usepackage[margin=1in]{geometry}
\usepackage{times}
\usepackage{amsmath}
\usepackage{amssymb}
\usepackage{enumitem}
\usepackage[table]{xcolor}
\usepackage{multirow}
\usepackage{booktabs}
\usepackage{longtable}
\usepackage{array}
\usepackage{graphicx}
\usepackage[authoryear,round]{natbib}
\usepackage{hyperref}
\definecolor{axisblue}{HTML}{D6E9F8}

\title{Can Activation Steering Substitute for Prompt-Based Affective Priming in LLM Digital Twins?}
\author{Junyi Fan \\ University of Southern California}
\date{}

\begin{document}

\maketitle
\begin{abstract}
Affective priming -- asking participants to write about a recalled event, such as a time they felt grateful or lacked control, and then measuring the induced state and its spillover onto related judgments -- is a standard way to manipulate emotion in the social sciences, and it also shifts the responses of LLM-based digital twins. The writing task, however, requires generating a full text. We test whether activation steering can replace it. Using Llama-3.1-8B-Instruct on 100 personas in two domains (gratitude, lack of control), we compare a real-prompt path replicating the original Twin-2K-500 Mega Study (J2) with an injection path that skips the writing and adds a mean-difference direction, extracted from human writing, to the residual stream at varying strengths. The real-prompt effects fall between the original human and digital-twin effects in 5 of 6 comparisons. Injection shifts the manipulation check in a dose-dependent way and matches the real-prompt effect at one strength in the gratitude domain ($\beta\approx2.7$), but overshoots beyond it, on the proximal measure as well, and asymmetrically across the two directions. A structure-matched No-Injection baseline separates two sources of difference: omitting the writing task raises the proximal measure in both domains, whereas the injected direction itself moves a distal measure in the lack-of-control domain. Substitution is therefore measure-dependent: validity established for one way of inducing a state does not carry over uniformly to another.
\end{abstract}

\section{Introduction}

Social scientists often manipulate emotional states with affective priming: participants reflect on and write about a recent event, and researchers test whether the induced state carries over to the responses that follow. A \emph{prime} prompt asks for an event tied to the target state and a \emph{baseline} prompt for a neutral or opposite event. A \emph{manipulation check} confirms that the state was induced, and \emph{proximal} measures (closely related to the target state) and \emph{distal} measures (more distantly related) show how far it spills over. In the gratitude version, participants recall a specific time they felt sincerely grateful for someone's kindness or help, or, at baseline, their morning routine \citep{emmons2003counting, desteno2014gratitude, oguni2024gratitude}. In the lack-of-control version, they describe an incident in which they had no control over the situation, or, at baseline, one in which they were in control \citep{bukowski2024control, chen2017control, lembregts2018numbers, whitson2008lacking}. ``Priming'' here refers to this recall-based induction of a state, not to the brief stimulus-presentation paradigms of cognitive psychology.

Whether such manipulations work on LLM-based stand-ins for people tests whether a simulated respondent reproduces not only static traits but also state-level shifts. Persona-conditioned models can reproduce subpopulation-level survey responses \citep{argyle2023outofone}, and ``digital twins'' built from rich personal data predict individuals' held-out answers with high accuracy \citep{park2024generative1000, toubia2025twin2k}. The J2 sub-study of the Twin-2K-500 Mega Study \citep{peng2026funhouse} asked whether affective priming works with digital twins. One thousand human participants and their twins were assigned to a prime or baseline writing condition in the gratitude or the lack-of-control domain, and then completed a manipulation check, a proximal measure (Elevation; Fatigue) and a distal measure (an adapted Empathic Concern scale; Desire for Predictability). Writing about gratitude or lack of control induced the intended states in twins, twins responded more strongly than humans, and proximal measures showed spillover while distal measures showed small or inconsistent effects. This over-responsiveness is one of several systematic distortions documented there, alongside cautionary work on LLM surrogates \citep{gao2025takecaution, hu2024persona, li2025persona} showing that incidental pipeline choices can change simulated respondents' behavior in ways unrelated to the manipulation. This raises a question for any alternative way of constructing a simulated state: does validity established for one method carry over to another?

The writing task is an obvious candidate for replacement. It requires the model to generate a complete piece of writing, which adds inference cost and leaves open whether the effect comes from the content the writing activates or from the act of writing itself. Activation steering offers an alternative. If concepts are approximately linear directions in activation space \citep{park2024linear}, a direction computed from contrastive examples can be added to hidden states at inference time without changing the input \citep{turner2023actadd, rimsky2024caa, zou2023repe}. The closest precedent, PsySET \citep{banayeeanzade2025psyset}, evaluates steering of emotions and personality and finds finer intensity control than prompting at some cost to output quality, with narrow mid-layer windows working best. Such evaluations score the steered model's own emotional expression. Whether injection reproduces the downstream consequences of an established human-subjects manipulation has, to our knowledge, not been tested.

We implement two parallel pipelines on one open-weight model (Llama-3.1-8B-Instruct; \citealp{grattafiori2024llama3}), using the J2 materials and instruments: a \textbf{real-prompt path}, in which the model completes the writing task before answering, and an \textbf{injection path}, in which the writing is skipped and a direction extracted from the real human writing is added at strength $\beta$. The real-prompt path is compared with the original study's human and digital-twin (GPT-4.1) results to provide an external calibration; the two paths are then compared with each other on the same personas. We ask four research questions:
\begin{enumerate}[nosep]
  \item[\textbf{RQ1}] How does the real-prompt effect of our small open-weight model compare with the human and digital-twin results of the original study?
  \item[\textbf{RQ2}] Can injection induce a measurable, dose-dependent shift in the manipulation check?
  \item[\textbf{RQ3}] Is there an injection strength at which injection and the real prompt produce comparable effects?
  \item[\textbf{RQ4}] Does injection reproduce the proximal--distal and valence structure of real-prompt priming?
\end{enumerate}

\paragraph{Contributions.} (i) A direct comparison of prompt-based and activation-injection induction of the same state in LLM digital twins on a published human-subjects paradigm, with an external calibration against the original human and twin results. (ii) Evidence that validity does not transport uniformly: injection is dose-dependent and can be matched to the real prompt on the manipulation check, but overshoots beyond the matching strength, on the proximal measure too, and asymmetrically across directions. (iii) A structure-matched No-Injection baseline that separates the effect of omitting the writing task from the effect of the injected vector. Code, generation logs and analysis scripts are released at \url{https://github.com/JunyiTim/twin-activation-injection}.

\section{Method}

\subsection{Setup}
\label{sec:setup}

\paragraph{Model.} All experiments use Llama-3.1-8B-Instruct in bfloat16 for generation, direction extraction and injection. Generation uses sampling at temperature $0.7$ (top-$p=0.9$, top-$k=50$), no fixed seed, and at most 1500 (gratitude) or 5000 (lack of control) new tokens, set by the length of the question set.

\paragraph{Domains and materials.} We replicate the two J2 domains. In the gratitude domain the prime asks participants to recall a specific experience in which they felt sincerely grateful (QID46) and the baseline asks for their morning routine (QID47) \citep{emmons2003counting, desteno2014gratitude, oguni2024gratitude}. In the lack-of-control domain the prime asks for an incident in which they had no control over the situation (QID31) and the baseline for one in which they were in control (QID32) \citep{bukowski2024control, chen2017control, lembregts2018numbers, whitson2008lacking}. Writing prompts and all measure items are taken verbatim from the original instrument (Appendix~\ref{app:prompts}).

\paragraph{Personas.} Personas are the \texttt{persona\_summary} field of Twin-2K-500 \citep{toubia2025twin2k}, about 2,000 words of demographics, percentile-scored psychometric instruments and a few verbatim answers, fetched on demand through the public dataset API. We use the first 100 personas in the dataset's order, identical in both domains (Appendix~\ref{app:impl}). Every persona is run through every condition, whereas in the original study each participant was assigned to one condition; conditions are therefore matched on persona, which the analysis exploits. No new human data were collected, and all human and original-twin results quoted are published values.

\paragraph{Measures.} Each measure is the mean of its items on a 7-point scale. Following the original analysis code, Fatigue items 3, 7, 8 and Desire-for-Predictability items 1, 4, 5 are reverse-coded ($8-x$) before averaging, and the lack-of-control manipulation check is analyzed on the raw scale (higher = more control). The \emph{manipulation check} is three emotion-intensity items (gratitude) or three sense-of-control items (lack of control) \citep{greenaway2015we}. The \emph{proximal measure} is Elevation (6 items; \citealp{schnall2010elevation, walsh2022positive}) or Fatigue (9 items; \citealp{sedek1990cognitive, bukowski2024control}); the Fatigue items refer to ``recalling and describing the event'' and are only partly defined when no writing occurs. The \emph{distal measure} is an adapted Empathic Concern scale (2 items; \citealp{oliveira2021gratitude}) or Desire for Predictability (8 items; \citealp{webster1994closure, lembregts2018numbers}).

\subsection{Experimental Design}

In each domain, two paths run on the same personas (Table~\ref{tab:design-overview}). In the \textbf{real-prompt path}, a single generation contains the writing task, the manipulation check, the proximal and the distal measure, as in the original pipeline; it has a prime and a baseline condition. In the \textbf{injection path} the writing task is omitted and the model answers the same questions while a direction vector is added to its hidden states at strength $\beta$. A \emph{dose arm} injects along the priming direction at four strengths ($\beta>0$: Low $1.8$, Moderate $2.5$, High $3.2$, Extreme $3.9$); a single-point \emph{active-control arm} injects the opposite pole of the same direction ($\beta=-2.5$); and a \emph{No-Injection} condition ($\beta=0$) is run once per persona as the shared zero point. No-Injection has the same prompt and output structure as the injection conditions and differs from them only in the vector, which lets us separate the effect of the injected vector from the effect of changing the pipeline.

\begin{table}[htbp]
\centering
\renewcommand{\arraystretch}{1.2}
\begin{tabular}{@{} l l l @{}}
\toprule
\rowcolor{axisblue}
\textbf{Path} & \textbf{Condition} & \textbf{Strength ($\beta$)} \\
\midrule
Real-prompt & Prime / Control & --- \\
Injection & Dose Injection & Low 1.8, Moderate 2.5, High 3.2, Extreme 3.9 \\
Injection & Active-Control Injection & Moderate ($-2.5$) \\
Injection & No-Injection & 0 \\
\bottomrule
\end{tabular}
\caption{Experimental design, identical in both domains. All eight conditions are run for each of the 100 personas.}
\label{tab:design-overview}
\end{table}

\subsection{Vector Extraction and Injection}
\label{sec:extraction}

\begin{figure}[htbp]
\centering
\includegraphics[width=0.95\linewidth]{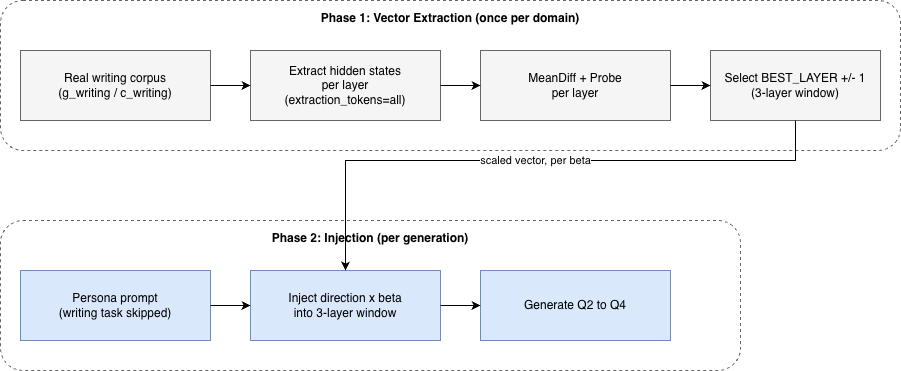}
\caption{Two-phase vector injection pipeline: direction vectors are extracted once per domain from real writing corpora (Phase 1), then injected into a small multi-layer window during generation, skipping the writing task entirely (Phase 2).}
\label{fig:injection-pipeline}
\end{figure}

The direction is extracted by mean difference (MeanDiff) from the human writing collected in the original Mega Study (506 examples for gratitude, 492 for lack of control), following activation addition \citep{turner2023actadd, rimsky2024caa, zou2023repe}; as in PsySET we inject into a small window of mid-depth layers (Figure~\ref{fig:injection-pipeline}). Each training example is a chat-formatted sequence (default system header, the writing instruction, the participant's writing). Let $\bar h_i^{(l)}\in\mathbb{R}^{4096}$ be the post-MLP residual-stream state at layer $l$ averaged over \emph{all} token positions of example $i$, and let $\mathcal{D}^{\pm}$ be the prime and baseline examples. Then
\begin{equation}
v^{(l)}=\frac{1}{|\mathcal{D}^+|}\sum_{i\in\mathcal{D}^+}\bar h_i^{(l)}-\frac{1}{|\mathcal{D}^-|}\sum_{i\in\mathcal{D}^-}\bar h_i^{(l)},\qquad \hat v^{(l)}=\frac{v^{(l)}}{\lVert v^{(l)}\rVert_2},
\end{equation}
and during generation the hook
\begin{equation}
\tilde h_t^{(l)}=h_t^{(l)}+\beta\,\hat v^{(l)},\qquad l\in\{L^*-1,L^*,L^*+1\},
\end{equation}
is applied at every token position (prompt and generated) of every forward pass, with each layer using its own direction. Because the average includes the instruction tokens, the direction mixes affective content, topic and instruction wording; this is deliberate, since the real manipulation conflates them too and we ask whether injection can substitute for the real manipulation, not whether it isolates a topic-free affect direction. A logistic-regression probe separated prime from baseline with held-out accuracy $1.000$ at every candidate layer (10--23) in both domains, so the probe could not discriminate among layers: $L^*=17$ is the midpoint of the candidate range and the window is $\{16,17,18\}$ (Appendix~\ref{app:method}). The four strengths were set after a three-persona pilot on the manipulation check (Appendix~\ref{app:method}); the matching strength estimated in RQ3 is therefore approximate.

\subsection{Parsing and Analysis}

An 8B model is less reliable than the original study's model at producing structured output, so free-form responses are parsed by an extractive LLM judge (\texttt{claude-sonnet-4-5}; \citealp{banayeeanzade2025psyset}) that returns the numeric answers or a failure marker, and parses with the wrong number of items are rejected. Failed generations are retried (at most two attempts in the gratitude domain, three in lack of control), every attempt is logged, and analyses use the final attempt of each cell when it succeeded (769 of 800 gratitude and 753 of 800 lack-of-control cells; Appendix~\ref{app:method}).

Because every persona appears in every condition, comparisons are paired by persona: paired $t$-tests on personas with successful generations in both conditions, with effect size $d_z$ (mean within-persona difference over its standard deviation; \citealp{cohen1988power}). The comparisons in the text address the research questions; Appendix~\ref{app:grid} reports all 39 comparisons per domain with Holm-adjusted $p$-values and Welch's test as a sensitivity check. Each injection condition is also compared with No-Injection (Holm over 15 within-path comparisons per domain). The strength at which the dose curve crosses the real-prompt effect is found by linear interpolation, with a 95\% percentile bootstrap interval (5,000 resamples of personas).

\section{Results}

All comparisons are paired by persona; $p$-values are uncorrected unless a Holm-adjusted value is given (Holm over the 39 comparisons per domain of Appendix~\ref{app:grid}); $d_z$ is the paired effect size. Per-condition means and standard errors are in Appendix~\ref{app:desc}.

\subsection{Calibration against the Original Study (RQ1)}

Table~\ref{tab:rq1-calibration} compares the priming effect on each measure in the real-prompt path with the published effects for the original study's human participants and digital twins.

\begin{table}[htbp]
\centering
\renewcommand{\arraystretch}{1.3}
\begin{tabular}{@{} l l c c c @{}}
\toprule
\rowcolor{axisblue}
\textbf{Domain} & \textbf{Measure} & \textbf{Human} & \textbf{This study} & \textbf{Digital twin} \\
\midrule
  \multirow{3}{*}{Gratitude} & Manipulation check & $0.81$ & $\mathbf{1.10}$ & $1.33$ \\
   & Proximal: Elevation & $0.79$ & $\mathbf{0.58}$ & $0.77$ \\
   & Distal: Empathic Concern & $-0.01$ & $\mathbf{0.02}$ & $0.24$ \\
\midrule
  \multirow{3}{*}{Lack of Control} & Manipulation check & $0.31$ & $\mathbf{0.63}$ & $1.00$ \\
   & Proximal: Fatigue & $0.18$ & $\mathbf{0.43}$ & $0.88$ \\
   & Distal: Desire for Predictability & $0.25$ & $\mathbf{0.13}$ & $0.05$ \\
\bottomrule
\end{tabular}
\caption{Priming effect (prime minus control) on each measure for this study's real-prompt path and for the original study's human participants and digital twins (published values). For the lack-of-control manipulation check, the effect is control minus prime on the raw scale (higher score = greater sense of control). Original-study values are differences between the published group means.}
\label{tab:rq1-calibration}
\end{table}

The effect lies between the human and the original digital-twin effect in 5 of 6 comparisons (the exception is gratitude Elevation, smaller than both). The small open-weight model therefore over-responds relative to humans on the manipulation checks and on the lack-of-control proximal measure, but less than the original larger, closed-source model; on the distal measures all effects are small in all three sources (at most 0.25). The comparison uses published group means and is descriptive; it serves as the benchmark for the injection results.

\subsection{Dose-Dependence of Injection (RQ2)}

\begin{figure}[htbp]
\centering
\includegraphics[width=\linewidth]{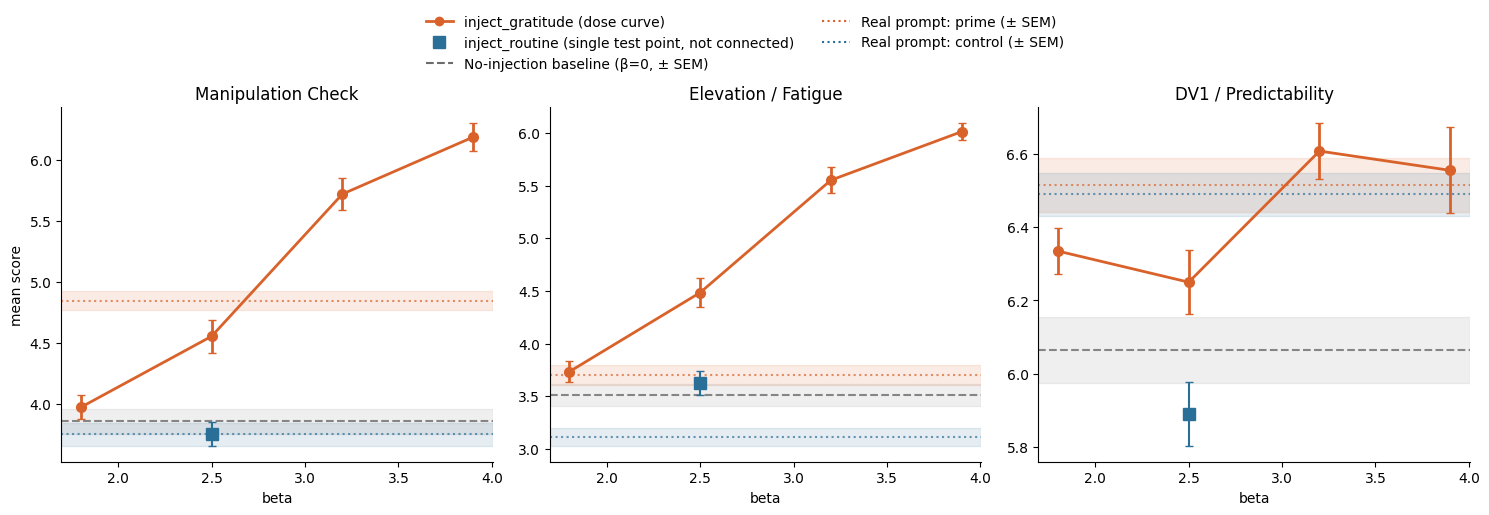}
\caption{Dose-response curves for the gratitude domain across the three measures. Dotted lines indicate the real-prompt reference conditions ($\pm$ SEM band), the dashed line the No-Injection baseline; the dose injection arm is a connected curve, the active-control injection arm an unconnected marker.}
\label{fig:dose-gratitude}
\end{figure}

\begin{figure}[htbp]
\centering
\includegraphics[width=\linewidth]{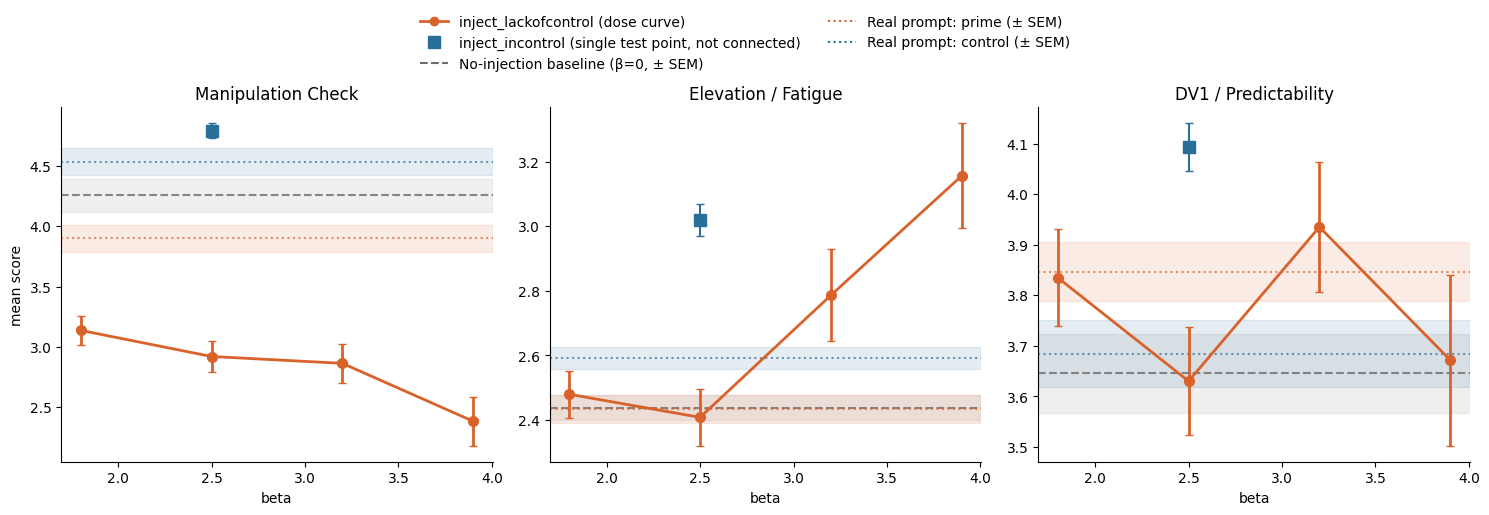}
\caption{Dose-response curves for the lack-of-control domain across the three measures, using the same conventions as Figure~\ref{fig:dose-gratitude}.}
\label{fig:dose-control}
\end{figure}

Figures~\ref{fig:dose-gratitude} and~\ref{fig:dose-control} show the dose-response patterns. The manipulation check responds monotonically to injection strength in both domains. In the gratitude domain, Low is already significantly below the real prime ($t(99)=-8.48$, $p<.0001$, $d_z=-0.85$) and Extreme exceeds it ($t(71)=10.20$, $p<.0001$, $d_z=1.20$). In the lack-of-control domain the pattern runs in the opposite direction, and Low already exceeds the real prime ($t(95)=-7.46$, $p<.0001$, $d_z=-0.76$). Injection thus induces a measurable, dose-dependent shift in the manipulation check (RQ2). Parsing success declines with strength, to 72\% (gratitude) and 77\% (lack of control) at Extreme (Appendix Table~\ref{tab:success-rate}), so results at Extreme are conditional on the generations that remained parseable.

\subsection{Matching Strength and Overshoot (RQ3)}

Table~\ref{tab:key-stats-dose} lists the key paired comparisons for RQ3 and RQ4.

\begin{table}[htbp]
\centering
\renewcommand{\arraystretch}{1.25}
\begin{tabular}{@{} >{\raggedright\arraybackslash}p{1.4cm} >{\raggedright\arraybackslash}p{4.1cm} l c c c @{}}
\toprule
\rowcolor{axisblue}
\textbf{Domain} & \textbf{Comparison} & \textbf{Measure} & \textbf{$d_z$} & \textbf{$t$ ($p$)} & \textbf{$p_{\text{Holm}}$} \\
\midrule
  \multirow{6}{1.4cm}{Gratitude} & Dose Injection (Low) vs.\ Prime & manip.\ check & $-0.85$ & $-8.48$ ($p<.0001$) & <.0001 \\
   & Dose Injection (Extreme) vs.\ Prime & manip.\ check & $1.20$ & $10.20$ ($p<.0001$) & <.0001 \\
   & Dose Injection (Low) vs.\ Prime & elevation & $0.04$ & $0.37$ ($p=.710$) & 1.000 \\
   & Dose Injection (Moderate) vs.\ Prime & elevation & $0.57$ & $5.68$ ($p<.0001$) & <.0001 \\
   & Prime vs.\ Control & elevation & $0.66$ & $6.58$ ($p<.0001$) & <.0001 \\
   & Prime vs.\ Control & dv1 & $0.03$ & $0.30$ ($p=.766$) & 1.000 \\
\midrule
  \multirow{5}{1.4cm}{Lack of Control} & Dose Injection (Low) vs.\ Prime & manip.\ check & $-0.76$ & $-7.46$ ($p<.0001$) & <.0001 \\
   & Dose Injection (Low) vs.\ Prime & fatigue & $1.03$ & $10.13$ ($p<.0001$) & <.0001 \\
   & Active-Control Injection (Moderate) vs.\ Control & manip.\ check & $0.28$ & $2.84$ ($p=.0056$) & .067 \\
   & Prime vs.\ Control & fatigue & $0.62$ & $6.20$ ($p<.0001$) & <.0001 \\
   & Prime vs.\ Control & dv1 & $0.13$ & $1.27$ ($p=.206$) & 1.000 \\
\bottomrule
\end{tabular}
\caption{Key paired comparisons for RQ3 (injection vs.\ real-prompt effect size) and the real-condition effects underlying RQ4. Differences are first condition minus second; $d_z$ is the mean within-persona difference divided by its standard deviation; $p_{\text{Holm}}$ is adjusted over the 39 comparisons per domain listed in Appendix~\ref{app:grid}.}
\label{tab:key-stats-dose}
\end{table}

In the gratitude domain the dose curve crosses the real prime mean on the manipulation check at $\beta\approx2.7$ (95\% bootstrap interval 2.5--2.8), between Moderate and High: it undershoots before and overshoots after. In the lack-of-control domain injection already exceeds the real-prompt effect at Low, so no matching point lies in the tested range. The overshoot is not confined to the manipulation check: on the proximal measure the dose arm exceeds the real prime from Moderate in the gratitude domain ($t(99)=5.68$, $p<.0001$, $d_z=0.57$) and from Low in the lack-of-control domain ($t(95)=10.13$, $p<.0001$, $d_z=1.03$). The two directions are not symmetric: the lack-of-control direction overshoots the manipulation check strongly ($d_z=-0.76$), whereas the in-control direction's effect at Moderate against the real control is much smaller ($t(98)=2.84$, $p=.0056$, $d_z=0.28$) and does not survive Holm adjustment ($p=.067$).

\subsection{Proximal--Distal Structure and Sources of Difference (RQ4)}
\label{sec:rq4}

\paragraph{Real-prompt path.} It reproduces the original proximal spillover in both domains: prime exceeds control on Elevation ($t(99)=6.58$, $p<.0001$, $d_z=0.66$) and on Fatigue ($t(98)=6.20$, $p<.0001$, $d_z=0.62$), while the distal measure does not differ (gratitude $d_z=0.03$, $p=.766$; lack of control $d_z=0.13$, $p=.206$), as in the original study.

\paragraph{Injection path against the real prompt.} The closest agreement is gratitude Elevation at Low, which does not differ significantly from the real prime ($t(99)=0.37$, $p=.710$, $d_z=0.04$; a non-significant difference is not evidence of equivalence, and we ran no equivalence test). In the lack-of-control domain injection does not match the real Fatigue effect at any strength (overshoot from Low, $d_z=1.03$).

\paragraph{Separating the pipeline from the vector.} Differences between injection-path and real-prompt conditions mix the injected vector with the change of pipeline (no writing task, shorter prompt, fewer questions, different output structure). Table~\ref{tab:pipeline-vector} separates them. The No-Injection baseline contains no vector, so its differences from the real control reflect the pipeline alone. It is higher on the proximal measure in both domains, with the same sign: Elevation in the gratitude domain ($t(99)=4.71$, $p<.0001$, $d_z=0.47$) and Fatigue in the lack-of-control domain ($t(96)=7.43$, $p<.0001$, $d_z=0.75$), where it is indistinguishable from the primed condition ($d_z=0.00$). The within-path contrast (active-control arm minus No-Injection, same prompt and output structure) isolates the vector. In the lack-of-control domain the in-control direction moves the manipulation check in the intended direction ($d_z=0.45$) and lowers the distal measure ($t(95)=-6.12$, $p<.0001$, $d_z=-0.63$), a measure on which the real manipulation has no reliable effect; the proximal measure does not change reliably. In the gratitude domain the active-control arm does not differ reliably from No-Injection on any measure, so its differences from the real conditions there come mostly from the pipeline. Within the injection path the dose arm raises the proximal measure strongly in both domains (gratitude, Moderate: $d_z=0.75$; lack of control, Low: $d_z=1.03$) and moves the distal measure only inconsistently across strengths (Appendix Table~\ref{tab:within-grid}). Persona-level agreement between conditions is moderate (Appendix~\ref{app:desc}), and we do not interpret it as individual-level fidelity.

\begin{table}[htbp]
\centering
\renewcommand{\arraystretch}{1.2}
\begin{tabular}{@{} >{\raggedright\arraybackslash}p{1.4cm} >{\raggedright\arraybackslash}p{5.2cm} l c c c @{}}
\toprule
\rowcolor{axisblue}
\textbf{Domain} & \textbf{Comparison} & \textbf{Measure} & \textbf{$d_z$} & \textbf{$t$ ($p$)} & \textbf{$p_{\text{Holm}}$} \\
\midrule
  \multirow{5}{1.4cm}{Gratitude} & No-Injection vs.\ Control & proximal & $0.47$ & $4.71$ ($p<.0001$) & .0001 \\
   & No-Injection vs.\ Control & distal & $-0.52$ & $-5.20$ ($p<.0001$) & <.0001 \\
   & Active-Control (Mod.) vs.\ No-Injection & manip.\ check & $-0.11$ & $-1.12$ ($p=.266$) & .620 \\
   & Active-Control (Mod.) vs.\ No-Injection & proximal & $0.12$ & $1.16$ ($p=.248$) & .620 \\
   & Active-Control (Mod.) vs.\ No-Injection & distal & $-0.17$ & $-1.73$ ($p=.087$) & .346 \\
\midrule
  \multirow{5}{1.4cm}{Lack of Control} & No-Injection vs.\ Control & proximal (fatigue) & $0.75$ & $7.43$ ($p<.0001$) & <.0001 \\
   & No-Injection vs.\ Prime & proximal (fatigue) & $0.00$ & $0.04$ ($p=.969$) & 1.000 \\
   & Active-Control (Mod.) vs.\ No-Injection & manip.\ check & $0.45$ & $4.36$ ($p<.0001$) & .0002 \\
   & Active-Control (Mod.) vs.\ No-Injection & proximal (fatigue) & $-0.12$ & $-1.21$ ($p=.228$) & .910 \\
   & Active-Control (Mod.) vs.\ No-Injection & distal & $-0.63$ & $-6.12$ ($p<.0001$) & <.0001 \\
\bottomrule
\end{tabular}
\caption{Pipeline versus vector. Comparisons of No-Injection with a real-prompt condition reflect the pipeline change alone; Active-Control (Moderate) minus No-Injection reflects the injected vector alone. $p_{\text{Holm}}$ is adjusted over the 39 comparisons per domain for rows against a real-prompt condition and over the 15 within-path comparisons for rows against No-Injection (Appendix~\ref{app:grid}). Conventions as in Table~\ref{tab:key-stats-dose}.}
\label{tab:pipeline-vector}
\end{table}

\section{Discussion}
\label{sec:discussion}

\paragraph{Summary.} The real-prompt path is a qualitatively faithful replication of the original manipulation: its effects fall between the human and original-twin effects in 5 of 6 comparisons and reproduce the proximal spillover and the null distal effect (RQ1). Injection induces a dose-dependent shift in the manipulation check (RQ2) that can be matched to the real-prompt effect at one strength in the gratitude domain ($\beta\approx2.7$) but overshoots beyond it, on the proximal measure as well, and asymmetrically across directions (RQ3). It reproduces the real proximal effect only at one strength in one domain and moves a distal measure where the real manipulation does not (RQ4).

\paragraph{Amplification, not reconstruction.} Twins over-respond to priming relative to humans in the original study, and our real-prompt path reproduced that at an intermediate magnitude. Injection does not reproduce this tendency through another channel; it exceeds it, on the manipulation check and on the proximal measure. This is more consistent with injection amplifying an existing property of the twin paradigm than with reconstructing the state real writing induces. The in-control direction complicates the picture: it has a small effect on its own manipulation check ($d_z=0.28$ against real control) yet lowers a distal measure within the injection path ($d_z=-0.63$ against No-Injection), which the real manipulation does not. One possibility is that the contrastive corpus mixes topical and affective signal asymmetrically across poles; the extraction also averaged over the instruction wording, which differs between poles, and the probe separated the classes perfectly at every layer, as expected when topic and wording are available. These observations are compatible with a direction that tracks topic and wording at least as much as affect.

\paragraph{Two sources of difference.} Because the injection path contains a structure-matched baseline, the vector and the pipeline can be separated (Table~\ref{tab:pipeline-vector}). Omitting the writing task raises the proximal measure relative to the real control in both domains regardless of the vector; the injected direction adds nothing reliable beyond No-Injection in the gratitude domain but lowers the distal measure in the lack-of-control domain. Which source dominates depends on domain and measure. The pipeline effect cannot be traced to one cause here: the injection path omits the writing, shortens the prompt, drops questions and changes the output structure, and the lack-of-control Fatigue items refer to an event that no longer exists. A neutral-writing control matched to the real-prompt structure would isolate the writing task; we leave it for future work.

\paragraph{Limitations.} We use one base model, one extraction method and a three-layer window that was not chosen by a discriminating criterion (the probe was at ceiling at every layer). The active-control arm was tested at a single strength. The strength grid was informed by a three-persona pilot whose personas are in the main sample, so the matching strength is approximate. We use paired tests rather than a single mixed-effects model and no equivalence tests, so ``no difference'' means non-significance. The retry limit differed between domains and results at Extreme are conditional on parseable generations. The direction was extracted from writing by the same participant pool whose twins we simulated, and we did not manually audit the parsed values. Finally, the paths differ in more than the vector, as discussed above.

\paragraph{Conclusion.} Activation steering can substitute for prompt-based affective priming on some measures, but not uniformly: on the manipulation check a well-chosen strength matches the real effect in one domain, whereas on the proximal and distal measures injection overshoots or shifts outcomes the real manipulation leaves unchanged, partly because of the vector and partly because of the changed pipeline. Researchers who use injection to induce states in digital twins should therefore calibrate strength separately for each outcome and include a no-injection control that matches the prompt and output format of the injection path. Next steps are a neutral-writing control, other models and extraction methods, and an extraction corpus that excludes the simulated personas.

\bibliographystyle{apalike}
\bibliography{references}

\begin{thebibliography}{}

\bibitem[Argyle et~al., 2023]{argyle2023outofone}
Argyle, L.~P., Busby, E.~C., Fulda, N., Gubler, J.~R., Rytting, C., and Wingate, D. (2023).
\newblock Out of one, many: Using language models to simulate human samples.
\newblock {\em Political Analysis}, 31(3):337--351.

\bibitem[Banayeeanzade et~al., 2025]{banayeeanzade2025psyset}
Banayeeanzade, A., Tak, A.~N., Bahrani, F., Bolourani, A., Blas, L., Ferrara, E., Gratch, J., and Karimireddy, S.~P. (2025).
\newblock Psychological steering in {LLMs}: An evaluation of effectiveness and trustworthiness.

\bibitem[Bukowski et~al., 2024]{bukowski2024control}
Bukowski, M., Potoczek, A., Barzykowski, K., Lautenbacher, J., and Inzlicht, M. (2024).
\newblock What do we manipulate when reminding people of (not) having control? in search of construct validity.
\newblock {\em Behavior Research Methods}, 56:3706--3724.

\bibitem[Chen et~al., 2017]{chen2017control}
Chen, C.~Y., Lee, L., and Yap, A.~J. (2017).
\newblock Control deprivation motivates acquisition of utilitarian products.
\newblock {\em Journal of Consumer Research}, 43(6):1031--1047.

\bibitem[Cohen, 1988]{cohen1988power}
Cohen, J. (1988).
\newblock {\em Statistical Power Analysis for the Behavioral Sciences}.
\newblock Lawrence Erlbaum Associates, Hillsdale, NJ, 2 edition.

\bibitem[DeSteno et~al., 2014]{desteno2014gratitude}
DeSteno, D., Li, Y., Dickens, L., and Lerner, J.~S. (2014).
\newblock Gratitude: A tool for reducing economic impatience.
\newblock {\em Psychological Science}, 25(6):1262--1267.

\bibitem[Emmons and McCullough, 2003]{emmons2003counting}
Emmons, R.~A. and McCullough, M.~E. (2003).
\newblock Counting blessings versus burdens: An experimental investigation of gratitude and subjective well-being in daily life.
\newblock {\em Journal of Personality and Social Psychology}, 84(2):377--389.

\bibitem[Gao et~al., 2025]{gao2025takecaution}
Gao, Y., Lee, D., Burtch, G., and Fazelpour, S. (2025).
\newblock Take caution in using {LLMs} as human surrogates.
\newblock {\em Proceedings of the National Academy of Sciences}, 122(24).
\newblock Preprint title: ``Take Caution in Using LLMs as Human Surrogates: Scylla Ex Machina'', arXiv:2410.19599.

\bibitem[Grattafiori et~al., 2024]{grattafiori2024llama3}
Grattafiori, A., Dubey, A., Jauhri, A., Pandey, A., Kadian, A., Al-Dahle, A., et~al. (2024).
\newblock The {Llama} 3 herd of models.

\bibitem[Greenaway et~al., 2015]{greenaway2015we}
Greenaway, K.~H., Haslam, S.~A., Cruwys, T., Branscombe, N.~R., Ysseldyk, R., and Heldreth, C. (2015).
\newblock From ``we'' to ``me'': Group identification enhances perceived personal control with consequences for health and well-being.
\newblock {\em Journal of Personality and Social Psychology}, 109(1):53--74.

\bibitem[Hu and Collier, 2024]{hu2024persona}
Hu, T. and Collier, N. (2024).
\newblock Quantifying the persona effect in {LLM} simulations.
\newblock In {\em Proceedings of the 62nd Annual Meeting of the Association for Computational Linguistics (Volume 1: Long Papers)}, pages 10289--10307.
\newblock arXiv:2402.10811.

\bibitem[Lembregts and Pandelaere, 2019]{lembregts2018numbers}
Lembregts, C. and Pandelaere, M. (2019).
\newblock Falling back on numbers: When preference for numerical product information increases after a personal control threat.
\newblock {\em Journal of Marketing Research}, 56(1):104--122.

\bibitem[Li et~al., 2025]{li2025persona}
Li, A., Chen, H., Namkoong, H., and Peng, T. (2025).
\newblock {LLM} generated persona is a promise with a catch.
\newblock In {\em Advances in Neural Information Processing Systems (NeurIPS), Position Paper Track}.
\newblock arXiv:2503.16527.

\bibitem[Oguni and Ishii, 2024]{oguni2024gratitude}
Oguni, R. and Ishii, C. (2024).
\newblock Gratitude promotes prosocial behavior even in uncertain situation.
\newblock {\em Scientific Reports}, 14:14379.

\bibitem[Oliveira et~al., 2021]{oliveira2021gratitude}
Oliveira, R., Bald{\'e}, A., Madeira, M., Ribeiro, T., and Arriaga, P. (2021).
\newblock The impact of writing about gratitude on the intention to engage in prosocial behaviors during the {COVID-19} outbreak.
\newblock {\em Frontiers in Psychology}, 12:588691.

\bibitem[Park et~al., 2024a]{park2024generative1000}
Park, J.~S., Zou, C.~Q., Shaw, A., Hill, B.~M., Cai, C., Morris, M.~R., Willer, R., Liang, P., and Bernstein, M.~S. (2024a).
\newblock Generative agent simulations of 1,000 people.

\bibitem[Park et~al., 2024b]{park2024linear}
Park, K., Choe, Y.~J., and Veitch, V. (2024b).
\newblock The linear representation hypothesis and the geometry of large language models.
\newblock In {\em Proceedings of the 41st International Conference on Machine Learning (ICML)}.
\newblock arXiv:2311.03658.

\bibitem[Peng et~al., 2026]{peng2026funhouse}
Peng, T., Brucks, M., Gui, G., Merlau, D.~J., Fan, G.~J., Ben~Sliman, M., Johnson, E.~J., Althenayyan, A., Bellezza, S., Donati, D., Fong, H., Friedman, E., Guevara, A., Hussein, M., Jerath, K., Kogut, B., Kumar, A., Lane, K., Li, H., Morwitz, V., Netzer, O., Perkowski, P., and Toubia, O. (2026).
\newblock Digital twins are funhouse mirrors: Five systematic distortions.
\newblock {\em Science Advances}, 12(36):eaeh8260.
\newblock Preprint: arXiv:2509.19088 (titled ``Digital Twins as Funhouse Mirrors: Five Key Distortions'').

\bibitem[Rimsky et~al., 2024]{rimsky2024caa}
Rimsky, N., Gabrieli, N., Schulz, J., Tong, M., Hubinger, E., and Turner, A.~M. (2024).
\newblock Steering {Llama} 2 via contrastive activation addition.
\newblock In {\em Proceedings of the 62nd Annual Meeting of the Association for Computational Linguistics (Volume 1: Long Papers)}, pages 15504--15522.
\newblock arXiv:2312.06681.

\bibitem[Schnall et~al., 2010]{schnall2010elevation}
Schnall, S., Roper, J., and Fessler, D. M.~T. (2010).
\newblock Elevation leads to altruistic behavior.
\newblock {\em Psychological Science}, 21(3):315--320.

\bibitem[Sedek and Kofta, 1990]{sedek1990cognitive}
Sedek, G. and Kofta, M. (1990).
\newblock When cognitive exertion does not yield cognitive gain: Toward an informational explanation of learned helplessness.
\newblock {\em Journal of Personality and Social Psychology}, 58(4):729--743.

\bibitem[Toubia et~al., 2025]{toubia2025twin2k}
Toubia, O., Gui, G.~Z., Peng, T., Merlau, D.~J., Li, A., and Chen, H. (2025).
\newblock Database report: {Twin-2K-500}: A data set for building digital twins of over 2,000 people based on their answers to over 500 questions.
\newblock {\em Marketing Science}, 44(6):1446--1455.
\newblock Preprint: arXiv:2505.17479.

\bibitem[Turner et~al., 2023]{turner2023actadd}
Turner, A.~M., Thiergart, L., Leech, G., Udell, D., Vazquez, J.~J., Mini, U., et~al. (2023).
\newblock Steering language models with activation engineering.
\newblock Earlier title: ``Activation Addition: Steering Language Models Without Optimization''.

\bibitem[Walsh et~al., 2022]{walsh2022positive}
Walsh, L.~C., Armenta, C.~N., Itzchakov, G., Fritz, M.~M., and Lyubomirsky, S. (2022).
\newblock More than merely positive: The immediate affective and motivational consequences of gratitude.
\newblock {\em Sustainability}, 14(14):8679.

\bibitem[Webster and Kruglanski, 1994]{webster1994closure}
Webster, D.~M. and Kruglanski, A.~W. (1994).
\newblock Individual differences in need for cognitive closure.
\newblock {\em Journal of Personality and Social Psychology}, 67(6):1049--1062.

\bibitem[Welch, 1947]{welch1947generalization}
Welch, B.~L. (1947).
\newblock The generalization of ``{Student's}'' problem when several different population variances are involved.
\newblock {\em Biometrika}, 34(1--2):28--35.

\bibitem[Whitson and Galinsky, 2008]{whitson2008lacking}
Whitson, J.~A. and Galinsky, A.~D. (2008).
\newblock Lacking control increases illusory pattern perception.
\newblock {\em Science}, 322(5898):115--117.

\bibitem[Zou et~al., 2023]{zou2023repe}
Zou, A., Phan, L., Chen, S., Campbell, J., Guo, P., Ren, R., et~al. (2023).
\newblock Representation engineering: A top-down approach to {AI} transparency.

\end{thebibliography}

\appendix

\section{Additional Method Details}
\label{app:method}

\subsection{Direction Extraction and Layer Selection}

\paragraph{Training examples.} Each example is a chat-formatted sequence consisting of the model's default system header, a user turn containing the writing instruction shown to the human participant (the prime or baseline prompt), and an assistant turn containing that participant's own writing. Write $x_i=(p_i,w_i)$ for instruction and writing, with token length $T_i$. Let $\mathcal{D}^+$ and $\mathcal{D}^-$ denote the prime- and baseline-condition examples, of sizes $n^+$ and $n^-$. In the gratitude domain, $n^+=249$ and $n^-=257$ (506 in total); in the lack-of-control domain, $n^+=256$ and $n^-=236$ (492 in total). Because the average over token positions (Section~\ref{sec:extraction}) runs over \emph{all} tokens of the sequence, including the instruction, the extracted direction mixes affective content, topical content, and the wording difference between the prime and baseline instructions; this is deliberate, since the real manipulation likewise conflates topic and affect, and our question is whether injection can substitute for the real manipulation rather than whether it isolates a topic-free affective direction.

\paragraph{Layer selection.} For each candidate layer $l\in\{10,\dots,23\}$ we train a logistic-regression probe (no intercept; regularization strength chosen on the training split, $C=100$ in the gratitude domain) to classify prime versus baseline from $\bar{h}_i^{(l)}$, using an 80/20 split (404 training and 102 test examples in the gratitude domain), and record held-out accuracy $\mathrm{Acc}(l)$. We define $L^{*}$ as the middle element of the set of layers attaining the maximum accuracy. In the gratitude domain, held-out accuracy was $1.000$ at \emph{every} candidate layer, so the probe did not discriminate among layers: $L^{*}=17$ is the midpoint of the candidate range rather than a data-driven maximum (the same held in the lack-of-control domain, again with $L^{*}=17$). The perfect separability itself is unsurprising given that the prime and baseline sequences differ in their instruction text and topic, and it should not be read as evidence that a purely affective signal has been isolated. Our choice of a mid-depth window is consistent with PsySET's finding that narrow mid-layer windows work best \citep{banayeeanzade2025psyset}.

\subsection{Choice of Injection Strengths}

The grid was set after a small pilot on the first three personas (which are also part of the main sample). At $\beta=2.0$ the mean manipulation-check score was $4.44$, below the level we were aiming for (about 6, near the top of the 7-point scale); at $\beta=4.5$ the first persona scored $7.00$ but generation then failed to produce parseable output and the sweep was halted. We therefore used four equally spaced strengths (step $0.7$) from $1.8$ to $3.9$, spanning the pilot range while staying below the strength at which generation broke down. The same grid was used in both domains. Because the grid was informed by the manipulation check, the location of the strength that matches the real-prompt effect (RQ3) should be read as approximate.

\subsection{Response Generation and Parsing}

An 8B instruction-tuned model is much less reliable than the larger model of the original study at producing well-formed structured output, and its free-form responses vary in format. Our prompts also omit the original pipeline's trailing JSON format instructions (the general instruction announces a format but none is appended), which further increases format variability. We therefore use an LLM-as-judge step, as in PsySET \citep{banayeeanzade2025psyset}: Claude (\texttt{claude-sonnet-4-5}, maximum 1024 output tokens, default sampling) receives each raw Llama response and either extracts the numeric answers into a fixed JSON schema or returns a failure marker if any item is missing, unclear, or unanswered. The judge is purely extractive; it is not asked to rate, repair, or complete responses. The code additionally rejects any parse whose item counts do not match the instrument (3, 6, and 2 items for the gratitude-domain Q2--Q4, and 3, 9, and 8 items for the lack-of-control domain). The full judge prompt is given in Appendix~\ref{app:prompts}. We did not manually audit the extracted values against the raw responses.

\begin{figure}[htbp]
\centering
\includegraphics[width=0.85\linewidth]{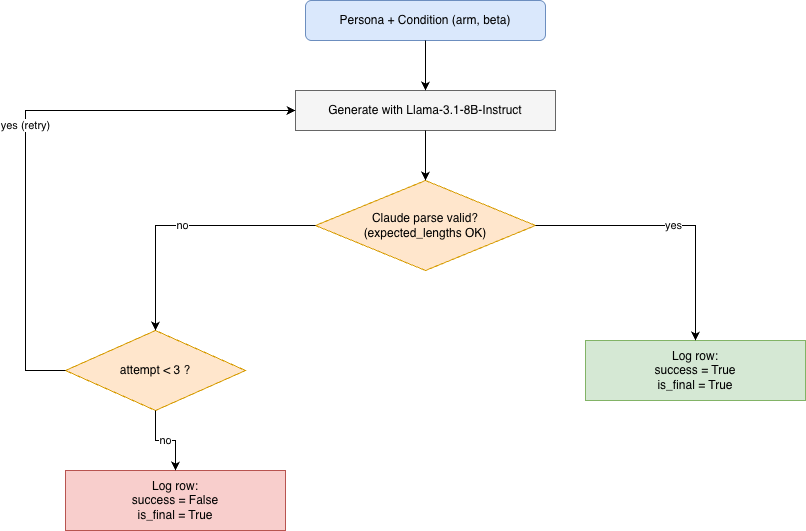}
\caption{Generation and LLM-as-judge parsing pipeline. Every attempt, including failed ones, is logged; each persona--condition cell receives at most two attempts (gratitude domain) or three (lack-of-control domain).}
\label{fig:gen-parse-flow}
\end{figure}

If a generation fails validation it is regenerated: each persona--condition cell receives at most two attempts in the gratitude domain (an initial attempt and one retry) and at most three in the lack-of-control domain, where the longer question set produced more parse failures. Every attempt, including failures, is logged with the raw Llama output, the judge's output, and the success flag. Figure~\ref{fig:gen-parse-flow} illustrates the pipeline.

\subsection{Data Quality and Sample Size}

Analyses use only the final attempt of each cell, and only when it succeeded (complete-case analysis). We verified that the success flag coincides exactly with whether valid numeric scores could be parsed (no flagged-successful row lacked parseable scores). The resulting sample size per condition is roughly 77--100, falling at high injection strength because of parse failures (reported per condition in Appendix~\ref{app:desc}), with a final failure rate below 10\% overall. Because failures concentrate at high strengths, results at Extreme are conditional on the generations that remained parseable.

\subsection{Statistical Analysis (Full Description)}

For each condition we compute the mean and standard error of each measure. Because every persona appears in every condition, comparisons are paired by persona: each injection condition is compared with the corresponding real-prompt condition, and the real-prompt prime and control conditions with each other, using paired $t$-tests on the personas whose final generations succeeded in both conditions (so the number of pairs is at most 100 and falls at high injection strength). We report $t$ with $n_{\text{pairs}}-1$ degrees of freedom, the paired effect size $d_z$ (the mean of the within-persona differences divided by their standard deviation; \citealp{cohen1988power}), and the correlation between the two conditions' scores. The comparisons in the main text were chosen to address the research questions, and the active-control analyses are exploratory. To guard against selective reporting, Appendix~\ref{app:grid} reports all 39 comparisons per domain (every injection condition, including No-Injection, against both real-prompt conditions on all three measures, plus prime versus control), with Holm-adjusted $p$-values computed over that full set. As a sensitivity check, the appendix also lists Welch's independent-samples $t$-test \citep{welch1947generalization}, which ignores the pairing and is therefore conservative. Because the injection path contains its own structure-matched baseline (No-Injection), we also compare each injection condition with No-Injection (paired, with Holm adjustment over these 15 within-path comparisons per domain, Appendix~\ref{app:grid}); this contrast removes any difference between the injection and real-prompt pipelines. The location of the injection strength that matches the real-prompt effect is estimated by linear interpolation of the dose curve, with a 95\% percentile bootstrap interval obtained by resampling personas (5{,}000 resamples). The real-prompt results are additionally compared descriptively with the digital-twin and human results published for the original study, forming the basis of the RQ1 analysis.

\section{Descriptive Results}
\label{app:desc}

\subsection{Condition Means}

Tables~\ref{tab:gratitude-full} and~\ref{tab:control-full} give the mean $\pm$ SEM of every measure in every condition (final successful generations). DV1 is the distal measure.

\begin{table}[htbp]
\centering
\renewcommand{\arraystretch}{1.3}
\begin{tabular}{@{} l c c c c @{}}
\toprule
\rowcolor{axisblue}
\textbf{Condition} & \textbf{n} & \textbf{Manip.\ Check} & \textbf{Elevation} & \textbf{DV1 (Concern)} \\
\midrule
Real Prime & 100 & $4.85 \pm 0.07$ & $3.70 \pm 0.09$ & $6.51 \pm 0.07$ \\
Real Control & 100 & $3.75 \pm 0.09$ & $3.12 \pm 0.09$ & $6.49 \pm 0.06$ \\
No-Injection Baseline & 100 & $3.85 \pm 0.10$ & $3.51 \pm 0.11$ & $6.07 \pm 0.09$ \\
Dose Injection (Low) & 100 & $3.97 \pm 0.10$ & $3.73 \pm 0.10$ & $6.33 \pm 0.06$ \\
Dose Injection (Moderate) & 100 & $4.55 \pm 0.14$ & $4.48 \pm 0.14$ & $6.25 \pm 0.09$ \\
Dose Injection (High) & 97 & $5.72 \pm 0.13$ & $5.55 \pm 0.13$ & $6.61 \pm 0.08$ \\
Dose Injection (Extreme) & 72 & $6.19 \pm 0.11$ & $6.01 \pm 0.08$ & $6.56 \pm 0.12$ \\
Active-Control Injection (Moderate) & 100 & $3.75 \pm 0.10$ & $3.62 \pm 0.11$ & $5.89 \pm 0.09$ \\
\bottomrule
\end{tabular}
\caption{Gratitude domain: mean $\pm$ SEM by condition (final successful generations; reverse-coding as described in Section~\ref{sec:setup}).}
\label{tab:gratitude-full}
\end{table}

\begin{table}[htbp]
\centering
\renewcommand{\arraystretch}{1.3}
\begin{tabular}{@{} l c c c c @{}}
\toprule
\rowcolor{axisblue}
\textbf{Condition} & \textbf{n} & \textbf{Manip.\ Check} & \textbf{Fatigue} & \textbf{DV1 (Predictability)} \\
\midrule
Real Prime & 99 & $3.90 \pm 0.11$ & $2.64 \pm 0.08$ & $5.50 \pm 0.07$ \\
Real Control & 100 & $4.53 \pm 0.11$ & $2.21 \pm 0.07$ & $5.38 \pm 0.08$ \\
No-Injection Baseline & 97 & $4.25 \pm 0.14$ & $2.65 \pm 0.08$ & $5.29 \pm 0.08$ \\
Dose Injection (Low) & 97 & $3.14 \pm 0.12$ & $3.59 \pm 0.11$ & $5.28 \pm 0.08$ \\
Dose Injection (Moderate) & 93 & $2.92 \pm 0.13$ & $3.73 \pm 0.11$ & $5.22 \pm 0.09$ \\
Dose Injection (High) & 91 & $2.86 \pm 0.16$ & $4.06 \pm 0.14$ & $5.20 \pm 0.07$ \\
Dose Injection (Extreme) & 77 & $2.39 \pm 0.20$ & $4.49 \pm 0.16$ & $4.75 \pm 0.13$ \\
Active-Control Injection (Moderate) & 99 & $4.79 \pm 0.06$ & $2.55 \pm 0.05$ & $4.73 \pm 0.05$ \\
\bottomrule
\end{tabular}
\caption{Lack-of-control domain: mean $\pm$ SEM by condition. Fatigue items 3, 7, 8 and Desire-for-Predictability items 1, 4, 5 are reverse-coded.}
\label{tab:control-full}
\end{table}

\subsection{Parsing Success Rate}

Table~\ref{tab:success-rate} reports the parsing success rate for the dose injection arm in both domains, by strength.

\begin{table}[htbp]
\centering
\renewcommand{\arraystretch}{1.3}
\begin{tabular}{@{} l c c c c @{}}
\toprule
\rowcolor{axisblue}
\textbf{Domain} & \textbf{Low} & \textbf{Moderate} & \textbf{High} & \textbf{Extreme} \\
\midrule
Gratitude & 100\% & 100\% & 97\% & \textbf{72\%} \\
Lack of Control & 97\% & 93\% & 91\% & \textbf{77\%} \\
\bottomrule
\end{tabular}
\caption{Parsing success rate for the dose injection arm, by domain and injection strength.}
\label{tab:success-rate}
\end{table}

Success rate declines monotonically with injection strength in both domains, dropping sharply at Extreme. Because this decline reflects only whether a generation could be structurally parsed, it is an outcome-independent, process-level indicator of over-injection: as strength increases, generation becomes less stable regardless of any particular downstream measure. It parallels the widening standard errors at High and Extreme (Tables~\ref{tab:gratitude-full} and~\ref{tab:control-full}), which suggests a common cause -- injection strength pushing the model outside its stable operating regime. Results at Extreme are conditional on the generations that remained parseable. By contrast, the active-control injection arm has among the \emph{highest} success rates in the dataset: 100\% at both tested strengths in the gratitude domain, and 97\% (No-Injection) and 99\% (Moderate) in the lack-of-control domain, so shifts in that arm cannot be attributed to selective attrition.

\subsection{Persona-Level Agreement}

Paired scores are moderately correlated across conditions. Between real prime and real control, $r$ on the proximal measure is 0.50 (gratitude) and 0.60 (lack of control). Between the dose arm at Low and real prime on the manipulation check, $r$ is 0.32 (gratitude) and 0.63 (lack of control). We have no human or twin benchmark for these values and do not interpret them as evidence of individual-level fidelity.

\section{Implementation Details}
\label{app:impl}

\paragraph{Software and hardware.} Experiments were run in Google Colab on a single GPU; the runtime was rebuilt between runs, so exact library versions may differ slightly across runs. Steering used the PsySET code base \citep{banayeeanzade2025psyset} (modified Llama implementation exposing a \texttt{hook\_after\_mlp\_hs} hook point after each decoder layer's MLP residual addition). Versions in the last recorded environment: \texttt{transformers} 4.56.1, \texttt{torch} 2.11.0 (CUDA 12.8).

\paragraph{Extraction pipeline.} Hidden states were extracted for candidate layers 10--23 (14 layers) in batches of 8, averaged over all token positions of each chat-formatted example, in bfloat16 and cast to float32 for the probe. Gratitude domain: 506 examples (249 prime, 257 baseline); probes were trained on 404 examples and evaluated on 102. Held-out probe accuracy was $1.000$ at every candidate layer in both domains, giving $L^{*}=17$ and an injection window of layers $\{16,17,18\}$ in both. Lack-of-control domain: 492 examples (256 prime, 236 baseline), i.e.\ 393 training and 99 test examples under the same split rule. As a reference point for how much of this separability is lexical, a bag-of-words (TF-IDF) logistic-regression classifier trained on the participants' writing alone (no instruction text; 20 random stratified 80/20 splits) reached held-out accuracy $1.000$ in the gratitude domain and $0.77$ (range $0.69$--$0.89$) in the lack-of-control domain. The activation probe is not directly comparable, since its input also contains the instruction wording, but the gratitude result shows that the two sets of writings are already perfectly separable by vocabulary alone.

\paragraph{Generation.} Ancestral sampling at temperature $0.7$; maximum new tokens 1500 (gratitude) and 5000 (lack of control); no fixed seed. Injection was implemented as forward hooks adding $\beta\hat v^{(l)}$ to the output of \texttt{hook\_after\_mlp\_hs} at layers 16--18, active for the entire call to \texttt{generate} (prompt prefill and every decoding step).

\paragraph{Pilot.} $\beta\in\{2.0,\,4.5\}$ on three personas: mean manipulation check $4.44$ ($n=3$) at $\beta=2.0$; $7.00$ for the first persona at $\beta=4.5$, followed by a parse failure, after which the sweep stopped.

\paragraph{Personas.} The first 100 personas of the \texttt{full\_persona} configuration of the Twin-2K-500 dataset (field \texttt{persona\_summary}), in dataset order. The first three pids are 574, 2001, and 1710. The full list, in run order, is: 574, 2001, 1710, 1277, 789, 1239, 820, 1913, 490, 1393, 1409, 1120, 64, 344, 623, 1447, 1014, 1869, 270, 717, 1573, 1827, 759, 394, 1497, 914, 169, 440, 1624, 1343, 127, 771, 1515, 896, 1072, 216, 1841, 258, 695, 972, 1196, 468, 1325, 141, 426, 1642, 1211, 808, 1891, 512, 1776, 288, 846, 1738, 2029, 191, 1975, 1108, 1692, 645, 1421, 1146, 322, 814, 1597, 294, 1243, 1724, 2035, 540, 1473, 617, 370, 50, 1969, 1114, 659, 1927, 1509, 1293, 590, 1813, 1547, 723, 244, 1020, 113, 689, 1377, 1610, 474, 920, 1339, 80, 1792, 1008, 1875, 222, 1046, 1521.

\paragraph{Data and code availability.} Code, the two final generation logs (880 rows for gratitude and 1,109 for lack of control, one row per attempt, including the raw Llama output and the judge's output), and the analysis scripts are available at \url{https://github.com/JunyiTim/twin-activation-injection}. Persona text is not redistributed; it can be fetched from the public Twin-2K-500 API as described above.

\section{Prompts}
\label{app:prompts}

\paragraph{System instruction (identical to the original Mega Study pipeline)}
\begin{footnotesize}
\begin{verbatim}
You are an AI assistant. Your task is to answer the 'New Survey Question' as if you are the 
person described in the 'Persona Profile' (which consists of their past survey responses). 
Remain consistent with the persona's previous answers and stated traits. Simulate their 
responses to new questions while accounting for human cognitive limitations, uncertainty, 
and biases.
Follow all instructions provided for the new question carefully regarding the format of your
 answer.
\end{verbatim}
\end{footnotesize}

\paragraph{User prompt structure.} The user turn is the concatenation of the header, the persona summary, the separator, the general instruction, and the question block, in this order; the original pipeline's trailing JSON format-instruction block is not used. The real-prompt path's question block is Q1 (writing task) followed by Q2--Q4; the injection path's is Q2--Q4 only. All text below is verbatim from the notebooks that produced the data.

\paragraph{Header, separator, and general instruction}
\begin{footnotesize}
\begin{verbatim}
HEADER:
## Persona Profile (This individual's past survey responses):

SEPARATOR:


---
## New Survey Question & Instructions (Please respond as the persona described above):

GENERAL INSTRUCTION:
Please answer the following questions as if you were taking this survey. The expected output
 is a JSON object and the format is provided in the end. 
---
\end{verbatim}
\end{footnotesize}

\paragraph{Writing prompts (gratitude domain)}
\begin{footnotesize}
\begin{verbatim}
QID46 (prime):
Please recall carefully and in detail a specific experience in the past when you felt 
sincerely grateful for someone's kindness or help. Spend two minutes detailing this 
experience below:

QID47 (baseline):
Please recall carefully and in detail the sequence of your morning routine (e.g., brushing 
teeth, changing clothes). Spend two minutes detailing this experience below:
\end{verbatim}
\end{footnotesize}

\paragraph{Writing prompts (lack-of-control domain)}
\begin{footnotesize}
\begin{verbatim}
QID31 (prime):
Please recall a particular incident in which something happened and you did not have control
 over the situation. Please describe the situation in which you felt a lack of control – 
what happened, how you felt, etc. Spend two minutes detailing this experience below:

QID32 (baseline):
Please recall a particular incident in which something happened and you were in control of 
the situation. Please describe the situation in which you felt in control – what happened, 
how you felt, etc. Spend two minutes detailing this experience below:
\end{verbatim}
\end{footnotesize}

\paragraph{Writing-task block (Q1) template}
\begin{footnotesize}
\begin{verbatim}
{instruction}
Question Type: Text Entry
Answer: [Masked]
\end{verbatim}
\end{footnotesize}

\paragraph{Question blocks (gratitude domain)}
\begin{footnotesize}
\begin{verbatim}
[Q_MANIP_CHECK]
Please indicate the intensity with which you feel each of the following emotions right now.
Question Type: Matrix
Options:
  1 = Not at all
  2 = 2
  3 = 3
  4 = 4
  5 = 5
  6 = 6
  7 = Extremely
1. Grateful
Answer: [Masked]
2. Thankful
Answer: [Masked]
3. Appreciative
Answer: [Masked]

[Q_ELEVATION]
Please indicate the extent to which you are feeling each of the below feelings right now.
Question Type: Matrix
Options:
  1 = Not at all
  2 = 2
  3 = 3
  4 = 4
  5 = 5
  6 = 6
  7 = Extremely
1. Feeling moved
Answer: [Masked]
2. Feeling uplifted
Answer: [Masked]
3. Feeling optimistic about humanity
Answer: [Masked]
4. Experiencing a warm feeling in the chest
Answer: [Masked]
5. Desiring to help others
Answer: [Masked]
6. Wanting to become a better person
Answer: [Masked]

[Q_DV1]
Please indicate how much you agree or disagree with the following statements regarding the 
immediate effects of tariffs on vulnerable communities:
Question Type: Matrix
Options:
  1 = Completely disagree
  2 = 2
  3 = 3
  4 = 4
  5 = 5
  6 = 6
  7 = Completely agree
1. I am very concerned about those most vulnerable to the effects of tariffs.
Answer: [Masked]
2. I feel compassion for those most affected by rising prices or job losses due to tariffs.
Answer: [Masked]
\end{verbatim}
\end{footnotesize}

\paragraph{Question blocks (lack-of-control domain)}
\begin{footnotesize}
\begin{verbatim}
[Q_MANIP_CHECK]
Please indicate the extent to which you agree with the following statement regarding how you
 feel right now.
Question Type: Matrix
Options:
  1 = Not at all
  2 = 2
  3 = 3
  4 = 4
  5 = 5
  6 = 6
  7 = Extremely
1. I feel in control of my life
Answer: [Masked]
2. I am free to live my life how I wish
Answer: [Masked]
3. My experiences in life are due to my own actions
Answer: [Masked]

[Q_FATIGUE]
Please indicate the extent to which you agree with the following statements relating to your
 experience reflecting on a recent event earlier in this survey.
Question Type: Matrix
Options:
  1 = Not at all
  2 = 2
  3 = 3
  4 = 4
  5 = 5
  6 = 6
  7 = Extremely
1. I felt discouraged when recalling and describing the event from my own past
Answer: [Masked]
2. My mind was running blank when recalling and describing the event from my own past
Answer: [Masked]
3. I was fully involved in recalling and describing the event
Answer: [Masked]
4. I found it very hard to think
Answer: [Masked]
5. It was difficult for me to get motivated to recall and describe the event
Answer: [Masked]
6. I was distracted and could not focus
Answer: [Masked]
7. Recalling and describing the event was a pleasant activity
Answer: [Masked]
8. The task of recalling and describing the event was very interesting
Answer: [Masked]
9. Very few ideas of a specific event came to my mind
Answer: [Masked]

[Q_PREDICTABILITY]
At this moment...
Question Type: Matrix
Options:
  1 = Completely disagree
  2 = 2
  3 = 3
  4 = 4
  5 = 5
  6 = 6
  7 = Completely agree
1. I would like to be around friends who are unpredictable
Answer: [Masked]
2. I would like to go to places where I have been before so that I know what to expect
Answer: [Masked]
3. I would not like to go into a situation without knowing what I can expect from it
Answer: [Masked]
4. I feel that it would be fun to change my plans at the last moment
Answer: [Masked]
5. I would enjoy the uncertainty of going into a new situation without knowing what might 
happen
Answer: [Masked]
6. I would not like to be with people who are capable of unexpected actions
Answer: [Masked]
7. I would prefer to socialize with familiar friends because I know what to expect from them
Answer: [Masked]
8. I dislike unpredictable situations
Answer: [Masked]
\end{verbatim}
\end{footnotesize}

\paragraph{Judge prompts.} \texttt{\{q1\_line\}}, \texttt{\{q1\_key\}} and \texttt{\{raw\_text\}} are filled in programmatically; the \texttt{Q1} line and key are included only for the real-prompt path.

\paragraph{Judge prompt (gratitude domain)}
\begin{footnotesize}
\begin{verbatim}
You will be shown a response from another AI model that was asked to answer survey questions
 using a 1-7 rating scale.

Q2 (3 items, each rated 1-7): Grateful, Thankful, Appreciative
Q3 (6 items, each rated 1-7): Feeling moved, Feeling uplifted, Feeling optimistic about 
humanity, Experiencing a warm feeling in the chest, Desiring to help others, Wanting to 
become a better person
Q4 (2 items, each rated 1-7): concern about tariffs' effect on vulnerable communities, 
compassion for those affected by tariffs
{q1_line}

Read the response below and extract the answers. If it clearly answers every item, output 
ONLY this JSON (no other text):
{{"Q2": [n, n, n], "Q3": [n, n, n, n, n, n], "Q4": [n, n]{q1_key}}}

If any item is missing, unclear, or not actually answered, output exactly:
{{"failed": true}}

Response to parse:
---
{raw_text}
---
\end{verbatim}
\end{footnotesize}

\paragraph{Judge prompt (lack-of-control domain)}
\begin{footnotesize}
\begin{verbatim}
You will be shown a response from another AI model that was asked to answer survey questions
 using a 1-7 rating scale.

Q2 (3 items, each rated 1-7): feeling in control of my life, feeling free to live my life 
how I wish, feeling my experiences are due to my own actions
Q3 (9 items, each rated 1-7): felt discouraged, mind running blank, fully involved, hard to 
think, hard to get motivated, distracted, pleasant activity, interesting task, few ideas 
came to mind
Q4 (8 items, each rated 1-7): desire for predictability/familiarity vs. enjoying 
unpredictability, across 8 statements
{q1_line}

Read the response below and extract the answers. If it clearly answers every item, output 
ONLY this JSON (no other text):
{{"Q2": [n, n, n], "Q3": [n, n, n, n, n, n, n, n, n], "Q4": [n, n, n, n, n, n, n, 
n]{q1_key}}}

If any item is missing, unclear, or not actually answered, output exactly:
{{"failed": true}}

Response to parse:
---
{raw_text}
---
\end{verbatim}
\end{footnotesize}

\section{All Paired Comparisons}
\label{app:grid}

Table~\ref{tab:grid-gratitude} and Table~\ref{tab:grid-control} list every comparison of each injection condition (including No-Injection) with both real-prompt conditions on all three measures, plus real prime versus real control: 39 comparisons per domain. $\Delta$ is the mean within-persona difference (first minus second), $n$ the number of personas with successful final generations in both conditions, $r$ the correlation between the two conditions' scores, $p_{\text{Holm}}$ the Holm-adjusted paired $p$-value over the 39 comparisons of that domain, and $p_{\text{Welch}}$ the uncorrected $p$ of Welch's independent-samples test (a sensitivity check that ignores the pairing).

{\footnotesize
\setlength{\tabcolsep}{3pt}
\begin{longtable}{@{} l l r r r r r r r r @{}}
\caption{Gratitude domain: all paired comparisons.}\label{tab:grid-gratitude}\\
\toprule
\rowcolor{axisblue}
\textbf{Comparison} & \textbf{Measure} & $n$ & $\Delta$ & $d_z$ & $t$ & $p$ & $p_{\text{Holm}}$ & $p_{\text{Welch}}$ & $r$ \\
\midrule
\endfirsthead
\toprule
\rowcolor{axisblue}
\textbf{Comparison} & \textbf{Measure} & $n$ & $\Delta$ & $d_z$ & $t$ & $p$ & $p_{\text{Holm}}$ & $p_{\text{Welch}}$ & $r$ \\
\midrule
\endhead
\bottomrule
\endfoot
Prime vs.\ Control & manip.\ check & 100 & $1.10$ & $1.10$ & $10.98$ & <.0001 & <.0001 & <.0001 & $0.29$ \\
Active-Ctrl (Moderate) vs.\ Prime & manip.\ check & 100 & $-1.10$ & $-1.16$ & $-11.55$ & <.0001 & <.0001 & <.0001 & $0.40$ \\
Active-Ctrl (Moderate) vs.\ Control & manip.\ check & 100 & $0.00$ & $0.00$ & $0.03$ & .972 & 1.000 & .980 & $0.48$ \\
Dose (Extreme) vs.\ Prime & manip.\ check & 72 & $1.33$ & $1.20$ & $10.20$ & <.0001 & <.0001 & <.0001 & $0.17$ \\
Dose (Extreme) vs.\ Control & manip.\ check & 72 & $2.43$ & $2.00$ & $16.95$ & <.0001 & <.0001 & <.0001 & $0.14$ \\
Dose (High) vs.\ Prime & manip.\ check & 97 & $0.86$ & $0.62$ & $6.07$ & <.0001 & <.0001 & <.0001 & $0.13$ \\
Dose (High) vs.\ Control & manip.\ check & 97 & $1.97$ & $1.31$ & $12.93$ & <.0001 & <.0001 & <.0001 & $0.10$ \\
Dose (Low) vs.\ Prime & manip.\ check & 100 & $-0.88$ & $-0.85$ & $-8.48$ & <.0001 & <.0001 & <.0001 & $0.32$ \\
Dose (Low) vs.\ Control & manip.\ check & 100 & $0.22$ & $0.23$ & $2.27$ & .025 & .327 & .100 & $0.47$ \\
Dose (Moderate) vs.\ Prime & manip.\ check & 100 & $-0.29$ & $-0.22$ & $-2.22$ & .029 & .349 & .063 & $0.34$ \\
Dose (Moderate) vs.\ Control & manip.\ check & 100 & $0.81$ & $0.68$ & $6.85$ & <.0001 & <.0001 & <.0001 & $0.53$ \\
No-Injection vs.\ Prime & manip.\ check & 100 & $-0.99$ & $-1.04$ & $-10.42$ & <.0001 & <.0001 & <.0001 & $0.45$ \\
No-Injection vs.\ Control & manip.\ check & 100 & $0.11$ & $0.14$ & $1.44$ & .154 & 1.000 & .436 & $0.71$ \\
Prime vs.\ Control & elevation & 100 & $0.58$ & $0.66$ & $6.58$ & <.0001 & <.0001 & <.0001 & $0.50$ \\
Active-Ctrl (Moderate) vs.\ Prime & elevation & 100 & $-0.08$ & $-0.08$ & $-0.83$ & .406 & 1.000 & .599 & $0.61$ \\
Active-Ctrl (Moderate) vs.\ Control & elevation & 100 & $0.51$ & $0.52$ & $5.18$ & <.0001 & <.0001 & .0004 & $0.54$ \\
Dose (Extreme) vs.\ Prime & elevation & 72 & $2.28$ & $2.07$ & $17.57$ & <.0001 & <.0001 & <.0001 & $0.12$ \\
Dose (Extreme) vs.\ Control & elevation & 72 & $2.88$ & $2.87$ & $24.37$ & <.0001 & <.0001 & <.0001 & $0.05$ \\
Dose (High) vs.\ Prime & elevation & 97 & $1.85$ & $1.46$ & $14.35$ & <.0001 & <.0001 & <.0001 & $0.35$ \\
Dose (High) vs.\ Control & elevation & 97 & $2.42$ & $1.77$ & $17.40$ & <.0001 & <.0001 & <.0001 & $0.19$ \\
Dose (Low) vs.\ Prime & elevation & 100 & $0.03$ & $0.04$ & $0.37$ & .710 & 1.000 & .806 & $0.57$ \\
Dose (Low) vs.\ Control & elevation & 100 & $0.62$ & $0.69$ & $6.88$ & <.0001 & <.0001 & <.0001 & $0.54$ \\
Dose (Moderate) vs.\ Prime & elevation & 100 & $0.78$ & $0.57$ & $5.68$ & <.0001 & <.0001 & <.0001 & $0.34$ \\
Dose (Moderate) vs.\ Control & elevation & 100 & $1.37$ & $1.03$ & $10.26$ & <.0001 & <.0001 & <.0001 & $0.36$ \\
No-Injection vs.\ Prime & elevation & 100 & $-0.19$ & $-0.21$ & $-2.12$ & .036 & .401 & .188 & $0.62$ \\
No-Injection vs.\ Control & elevation & 100 & $0.39$ & $0.47$ & $4.71$ & <.0001 & .0001 & .0046 & $0.65$ \\
Prime vs.\ Control & dv1 & 100 & $0.03$ & $0.03$ & $0.30$ & .766 & 1.000 & .791 & $0.22$ \\
Active-Ctrl (Moderate) vs.\ Prime & dv1 & 100 & $-0.62$ & $-0.68$ & $-6.84$ & <.0001 & <.0001 & <.0001 & $0.36$ \\
Active-Ctrl (Moderate) vs.\ Control & dv1 & 100 & $-0.60$ & $-0.81$ & $-8.12$ & <.0001 & <.0001 & <.0001 & $0.54$ \\
Dose (Extreme) vs.\ Prime & dv1 & 72 & $-0.02$ & $-0.02$ & $-0.16$ & .877 & 1.000 & .771 & $0.04$ \\
Dose (Extreme) vs.\ Control & dv1 & 72 & $-0.01$ & $-0.01$ & $-0.11$ & .911 & 1.000 & .619 & $0.18$ \\
Dose (High) vs.\ Prime & dv1 & 97 & $0.10$ & $0.10$ & $0.95$ & .345 & 1.000 & .378 & $0.07$ \\
Dose (High) vs.\ Control & dv1 & 97 & $0.10$ & $0.13$ & $1.31$ & .194 & 1.000 & .218 & $0.33$ \\
Dose (Low) vs.\ Prime & dv1 & 100 & $-0.18$ & $-0.21$ & $-2.11$ & .037 & .401 & .065 & $0.23$ \\
Dose (Low) vs.\ Control & dv1 & 100 & $-0.15$ & $-0.23$ & $-2.33$ & .022 & .306 & .074 & $0.40$ \\
Dose (Moderate) vs.\ Prime & dv1 & 100 & $-0.27$ & $-0.26$ & $-2.61$ & .011 & .158 & .022 & $0.22$ \\
Dose (Moderate) vs.\ Control & dv1 & 100 & $-0.24$ & $-0.31$ & $-3.14$ & .0022 & .035 & .024 & $0.52$ \\
No-Injection vs.\ Prime & dv1 & 100 & $-0.45$ & $-0.50$ & $-4.95$ & <.0001 & <.0001 & .0002 & $0.40$ \\
No-Injection vs.\ Control & dv1 & 100 & $-0.42$ & $-0.52$ & $-5.20$ & <.0001 & <.0001 & .0001 & $0.46$ \\
\end{longtable}
}

{\footnotesize
\setlength{\tabcolsep}{3pt}
\begin{longtable}{@{} l l r r r r r r r r @{}}
\caption{Lack-of-control domain: all paired comparisons.}\label{tab:grid-control}\\
\toprule
\rowcolor{axisblue}
\textbf{Comparison} & \textbf{Measure} & $n$ & $\Delta$ & $d_z$ & $t$ & $p$ & $p_{\text{Holm}}$ & $p_{\text{Welch}}$ & $r$ \\
\midrule
\endfirsthead
\toprule
\rowcolor{axisblue}
\textbf{Comparison} & \textbf{Measure} & $n$ & $\Delta$ & $d_z$ & $t$ & $p$ & $p_{\text{Holm}}$ & $p_{\text{Welch}}$ & $r$ \\
\midrule
\endhead
\bottomrule
\endfoot
Prime vs.\ Control & manip.\ check & 99 & $-0.63$ & $-0.65$ & $-6.42$ & <.0001 & <.0001 & <.0001 & $0.61$ \\
Active-Ctrl (Moderate) vs.\ Prime & manip.\ check & 98 & $0.88$ & $0.90$ & $8.95$ & <.0001 & <.0001 & <.0001 & $0.46$ \\
Active-Ctrl (Moderate) vs.\ Control & manip.\ check & 99 & $0.28$ & $0.28$ & $2.84$ & .0056 & .067 & .046 & $0.51$ \\
Dose (Extreme) vs.\ Prime & manip.\ check & 76 & $-1.59$ & $-0.75$ & $-6.57$ & <.0001 & <.0001 & <.0001 & $-0.01$ \\
Dose (Extreme) vs.\ Control & manip.\ check & 77 & $-2.24$ & $-1.10$ & $-9.67$ & <.0001 & <.0001 & <.0001 & $0.02$ \\
Dose (High) vs.\ Prime & manip.\ check & 91 & $-1.06$ & $-0.59$ & $-5.62$ & <.0001 & <.0001 & <.0001 & $0.13$ \\
Dose (High) vs.\ Control & manip.\ check & 91 & $-1.67$ & $-0.93$ & $-8.84$ & <.0001 & <.0001 & <.0001 & $0.14$ \\
Dose (Low) vs.\ Prime & manip.\ check & 96 & $-0.76$ & $-0.76$ & $-7.46$ & <.0001 & <.0001 & <.0001 & $0.63$ \\
Dose (Low) vs.\ Control & manip.\ check & 97 & $-1.40$ & $-1.30$ & $-12.85$ & <.0001 & <.0001 & <.0001 & $0.59$ \\
Dose (Moderate) vs.\ Prime & manip.\ check & 92 & $-1.02$ & $-0.88$ & $-8.45$ & <.0001 & <.0001 & <.0001 & $0.50$ \\
Dose (Moderate) vs.\ Control & manip.\ check & 93 & $-1.61$ & $-1.36$ & $-13.07$ & <.0001 & <.0001 & <.0001 & $0.49$ \\
No-Injection vs.\ Prime & manip.\ check & 96 & $0.37$ & $0.38$ & $3.70$ & .0004 & .0051 & .045 & $0.70$ \\
No-Injection vs.\ Control & manip.\ check & 97 & $-0.25$ & $-0.24$ & $-2.40$ & .018 & .184 & .120 & $0.66$ \\
Prime vs.\ Control & fatigue & 99 & $0.42$ & $0.62$ & $6.20$ & <.0001 & <.0001 & <.0001 & $0.60$ \\
Active-Ctrl (Moderate) vs.\ Prime & fatigue & 98 & $-0.09$ & $-0.12$ & $-1.15$ & .253 & 1.000 & .380 & $0.42$ \\
Active-Ctrl (Moderate) vs.\ Control & fatigue & 99 & $0.33$ & $0.50$ & $4.99$ & <.0001 & <.0001 & .0001 & $0.46$ \\
Dose (Extreme) vs.\ Prime & fatigue & 76 & $1.96$ & $1.41$ & $12.31$ & <.0001 & <.0001 & <.0001 & $0.21$ \\
Dose (Extreme) vs.\ Control & fatigue & 77 & $2.35$ & $1.78$ & $15.66$ & <.0001 & <.0001 & <.0001 & $0.28$ \\
Dose (High) vs.\ Prime & fatigue & 91 & $1.44$ & $1.24$ & $11.84$ & <.0001 & <.0001 & <.0001 & $0.46$ \\
Dose (High) vs.\ Control & fatigue & 91 & $1.82$ & $1.55$ & $14.83$ & <.0001 & <.0001 & <.0001 & $0.44$ \\
Dose (Low) vs.\ Prime & fatigue & 96 & $0.95$ & $1.03$ & $10.13$ & <.0001 & <.0001 & <.0001 & $0.53$ \\
Dose (Low) vs.\ Control & fatigue & 97 & $1.37$ & $1.51$ & $14.91$ & <.0001 & <.0001 & <.0001 & $0.52$ \\
Dose (Moderate) vs.\ Prime & fatigue & 92 & $1.11$ & $1.11$ & $10.66$ & <.0001 & <.0001 & <.0001 & $0.47$ \\
Dose (Moderate) vs.\ Control & fatigue & 93 & $1.53$ & $1.41$ & $13.59$ & <.0001 & <.0001 & <.0001 & $0.33$ \\
No-Injection vs.\ Prime & fatigue & 96 & $0.00$ & $0.00$ & $0.04$ & .969 & 1.000 & .896 & $0.71$ \\
No-Injection vs.\ Control & fatigue & 97 & $0.43$ & $0.75$ & $7.43$ & <.0001 & <.0001 & <.0001 & $0.70$ \\
Prime vs.\ Control & dv1 & 99 & $0.11$ & $0.13$ & $1.27$ & .206 & 1.000 & .238 & $0.30$ \\
Active-Ctrl (Moderate) vs.\ Prime & dv1 & 98 & $-0.77$ & $-1.03$ & $-10.21$ & <.0001 & <.0001 & <.0001 & $0.32$ \\
Active-Ctrl (Moderate) vs.\ Control & dv1 & 99 & $-0.64$ & $-0.90$ & $-8.91$ & <.0001 & <.0001 & <.0001 & $0.47$ \\
Dose (Extreme) vs.\ Prime & dv1 & 76 & $-0.69$ & $-0.55$ & $-4.79$ & <.0001 & .0001 & <.0001 & $0.10$ \\
Dose (Extreme) vs.\ Control & dv1 & 77 & $-0.59$ & $-0.44$ & $-3.84$ & .0003 & .0038 & <.0001 & $0.02$ \\
Dose (High) vs.\ Prime & dv1 & 91 & $-0.31$ & $-0.33$ & $-3.11$ & .0025 & .033 & .0038 & $0.09$ \\
Dose (High) vs.\ Control & dv1 & 91 & $-0.17$ & $-0.16$ & $-1.54$ & .128 & .894 & .108 & $0.02$ \\
Dose (Low) vs.\ Prime & dv1 & 96 & $-0.22$ & $-0.24$ & $-2.31$ & .023 & .210 & .035 & $0.20$ \\
Dose (Low) vs.\ Control & dv1 & 97 & $-0.09$ & $-0.09$ & $-0.91$ & .364 & 1.000 & .389 & $0.23$ \\
Dose (Moderate) vs.\ Prime & dv1 & 92 & $-0.26$ & $-0.29$ & $-2.79$ & .0064 & .070 & .012 & $0.34$ \\
Dose (Moderate) vs.\ Control & dv1 & 93 & $-0.15$ & $-0.15$ & $-1.45$ & .151 & .908 & .181 & $0.24$ \\
No-Injection vs.\ Prime & dv1 & 96 & $-0.18$ & $-0.22$ & $-2.14$ & .035 & .281 & .056 & $0.44$ \\
No-Injection vs.\ Control & dv1 & 97 & $-0.06$ & $-0.06$ & $-0.58$ & .562 & 1.000 & .451 & $0.33$ \\
\end{longtable}
}

Table~\ref{tab:within-grid} lists the within-path comparisons: each injection condition minus No-Injection, which share the same prompt and output structure (15 comparisons per domain, Holm-adjusted within domain).

{\footnotesize
\setlength{\tabcolsep}{3pt}
\begin{longtable}{@{} l l l r r r r r r @{}}
\caption{Within-path paired comparisons: injection condition minus No-Injection.}\label{tab:within-grid}\\
\toprule
\rowcolor{axisblue}
\textbf{Domain} & \textbf{Condition} & \textbf{Measure} & $n$ & $\Delta$ & $d_z$ & $t$ & $p$ & $p_{\text{Holm}}$ \\
\midrule
\endfirsthead
\toprule
\rowcolor{axisblue}
\textbf{Domain} & \textbf{Condition} & \textbf{Measure} & $n$ & $\Delta$ & $d_z$ & $t$ & $p$ & $p_{\text{Holm}}$ \\
\midrule
\endhead
\bottomrule
\endfoot
Gratitude & Dose (Low) & manip.\ check & 100 & $0.12$ & $0.13$ & $1.27$ & .207 & .620 \\
Gratitude & Dose (Low) & proximal & 100 & $0.22$ & $0.24$ & $2.39$ & .019 & .111 \\
Gratitude & Dose (Low) & distal & 100 & $0.27$ & $0.34$ & $3.39$ & .0010 & .0080 \\
Gratitude & Dose (Moderate) & manip.\ check & 100 & $0.70$ & $0.58$ & $5.81$ & <.0001 & <.0001 \\
Gratitude & Dose (Moderate) & proximal & 100 & $0.97$ & $0.75$ & $7.46$ & <.0001 & <.0001 \\
Gratitude & Dose (Moderate) & distal & 100 & $0.18$ & $0.19$ & $1.92$ & .058 & .288 \\
Gratitude & Dose (High) & manip.\ check & 97 & $1.85$ & $1.24$ & $12.17$ & <.0001 & <.0001 \\
Gratitude & Dose (High) & proximal & 97 & $2.03$ & $1.43$ & $14.09$ & <.0001 & <.0001 \\
Gratitude & Dose (High) & distal & 97 & $0.53$ & $0.57$ & $5.61$ & <.0001 & <.0001 \\
Gratitude & Dose (Extreme) & manip.\ check & 72 & $2.24$ & $1.71$ & $14.53$ & <.0001 & <.0001 \\
Gratitude & Dose (Extreme) & proximal & 72 & $2.35$ & $2.06$ & $17.47$ & <.0001 & <.0001 \\
Gratitude & Dose (Extreme) & distal & 72 & $0.41$ & $0.34$ & $2.91$ & .0049 & .034 \\
Gratitude & Active-Ctrl (Moderate) & manip.\ check & 100 & $-0.10$ & $-0.11$ & $-1.12$ & .266 & .620 \\
Gratitude & Active-Ctrl (Moderate) & proximal & 100 & $0.11$ & $0.12$ & $1.16$ & .248 & .620 \\
Gratitude & Active-Ctrl (Moderate) & distal & 100 & $-0.17$ & $-0.17$ & $-1.73$ & .087 & .346 \\
Lack of control & Dose (Low) & manip.\ check & 94 & $-1.16$ & $-0.98$ & $-9.52$ & <.0001 & <.0001 \\
Lack of control & Dose (Low) & proximal & 94 & $0.95$ & $1.03$ & $10.01$ & <.0001 & <.0001 \\
Lack of control & Dose (Low) & distal & 94 & $-0.01$ & $-0.01$ & $-0.05$ & .959 & 1.000 \\
Lack of control & Dose (Moderate) & manip.\ check & 90 & $-1.30$ & $-1.01$ & $-9.59$ & <.0001 & <.0001 \\
Lack of control & Dose (Moderate) & proximal & 90 & $1.09$ & $1.13$ & $10.68$ & <.0001 & <.0001 \\
Lack of control & Dose (Moderate) & distal & 90 & $-0.05$ & $-0.05$ & $-0.50$ & .616 & 1.000 \\
Lack of control & Dose (High) & manip.\ check & 88 & $-1.42$ & $-0.72$ & $-6.78$ & <.0001 & <.0001 \\
Lack of control & Dose (High) & proximal & 88 & $1.42$ & $1.22$ & $11.46$ & <.0001 & <.0001 \\
Lack of control & Dose (High) & distal & 88 & $-0.11$ & $-0.12$ & $-1.09$ & .279 & .910 \\
Lack of control & Dose (Extreme) & manip.\ check & 74 & $-2.05$ & $-0.92$ & $-7.92$ & <.0001 & <.0001 \\
Lack of control & Dose (Extreme) & proximal & 74 & $1.94$ & $1.48$ & $12.69$ & <.0001 & <.0001 \\
Lack of control & Dose (Extreme) & distal & 74 & $-0.46$ & $-0.34$ & $-2.95$ & .0042 & .021 \\
Lack of control & Active-Ctrl (Moderate) & manip.\ check & 96 & $0.53$ & $0.45$ & $4.36$ & <.0001 & .0002 \\
Lack of control & Active-Ctrl (Moderate) & proximal & 96 & $-0.08$ & $-0.12$ & $-1.21$ & .228 & .910 \\
Lack of control & Active-Ctrl (Moderate) & distal & 96 & $-0.55$ & $-0.63$ & $-6.12$ & <.0001 & <.0001 \\
\end{longtable}
}

\end{document}